\documentclass[11pt,letterpaper]{article}
\pdfoutput=1
\usepackage[hyphens]{url}
\usepackage[hyperref]{emnlp2017}
\usepackage{times}
\usepackage{latexsym}
\usepackage{times}
\usepackage{latexsym}
\usepackage{amsmath,amssymb}
\usepackage{multirow}
\usepackage{color}
\usepackage{graphicx}
\usepackage{bbm}
\usepackage{xspace}
\usepackage{wasysym}
\usepackage{algorithmic}
\usepackage{float}
\usepackage{mathtools}
\usepackage{pgfplots}
\pgfplotsset{compat=1.14}
\usepackage{placeins}
\usepackage{tabularx}
\usepgfplotslibrary{groupplots}
\emnlpfinalcopy

\newcolumntype{L}[1]{>{\raggedright\arraybackslash}p{#1}}
\newcolumntype{C}[1]{>{\centering\arraybackslash}p{#1}}
\newcolumntype{R}[1]{>{\raggedleft\arraybackslash}p{#1}}\mathtoolsset{showonlyrefs}

\newenvironment{itemizesquish}{\begin{list}{\labelitemi}{\setlength{\itemsep}{-0.2em}\setlength{\labelwidth}{0.5em}\setlength{\leftmargin}{\labelwidth}
\addtolength{\leftmargin}{\labelsep}}}{\end{list}}

\newcommand{\sent}{s}
\newcommand{\sentt}{s_t}

\newcommand{\paragramsl}{\textsc{paragram-sl999}\xspace}

\newcommand{\ganlong}{\textsc{gated recurrent averaging network}\xspace}
\newcommand{\gan}{\textsc{GRAN}\xspace}

\newcommand{\wordavg}{\textsc{Avg}\xspace}
\newcommand{\simpwiki}{SimpWiki\xspace}

\newcommand\norm[1]{\left\lVert#1\right\rVert}
\DeclareMathOperator*{\argmax}{argmax}

\title{Learning Paraphrastic Sentence Embeddings\\
from Back-Translated Bitext}

\author{John Wieting$^1$\ \ \ \ \ \ Jonathan Mallinson$^2$\ \ \ \ \ \ Kevin Gimpel$^1$\\
$^1$Toyota Technological Institute at Chicago, Chicago, IL, 60637, USA\\
$^2$School of Informatics, University of Edinburgh, Edinburgh, UK\\
\tt{\{jwieting,kgimpel\}@ttic.edu},  \tt{j.mallinson@ed.ac.uk}
}

\date{}

\begin{document}
\maketitle
\begin{abstract}
We consider the problem of learning general-purpose, paraphrastic sentence embeddings in the setting of \newcite{wieting-16-full}. 
We use 
neural machine translation to generate sentential paraphrases via back-translation of bilingual sentence pairs. We evaluate the paraphrase pairs by their ability to serve as training data for learning paraphrastic sentence embeddings. 
We find that the data quality is stronger than prior work based on bitext and on par with 
manually-written
English paraphrase pairs, with the advantage that our approach can scale up to generate large training sets for many languages and domains. 
We experiment with several language pairs and data sources, and develop 
a variety of data filtering techniques. 
In the process, we explore how 
neural machine translation output differs from human-written sentences, finding clear differences in length, 
the amount of repetition, and the use of rare words.\footnote{Generated paraphrases and code are  available at \url{http://ttic.uchicago.edu/~wieting}.}
\end{abstract}

\section{Introduction}
Pretrained word embeddings have received a great deal of attention from the research community, but there is much less work on developing pretrained embeddings for \emph{sentences}. Here we target sentence embeddings that are ``paraphrastic'' in the sense that two sentences with similar meanings are close in the embedding space. \newcite{wieting-16-full} developed paraphrastic sentence embeddings that are useful for semantic textual similarity tasks and can also be used as initialization for supervised semantic tasks. 

To learn their sentence embeddings, Wieting et al.~used 
the Paraphrase Database (PPDB) \cite{GanitkevitchDC13}. 
PPDB contains a large set of paraphrastic textual fragments extracted automatically from bilingual text (``bitext''), which is readily available for languages and domains. 
Versions of PPDB have been released for several languages~\cite{ganitkevitch2014multilingual}. 

However, more recent work has shown that the fragmental nature of PPDB's pairs can be problematic, especially for recurrent networks~\cite{wieting-17-full}. Better performance can be achieved with a smaller set of sentence pairs derived from aligning Simple English and standard English Wikipedia~\cite{coster2011simple}. While effective, this type of data is inherently limited in size and scope, and not available for languages other than English. 

PPDB is appealing in that it only requires bitext. 
We would like to retain this property but develop a data resource with sentence pairs rather than phrase pairs. 
We turn to neural machine translation (NMT) \cite{sutskever2014sequence,bahdanau2014neural,sennrich2015improving}, which has matured recently to yield strong performance especially in terms of producing grammatical outputs. 

\begin{table}[t]
\setlength{\tabcolsep}{3pt} 
\small
\centering
\begin{tabular} { | c  p{0.9\columnwidth} | } \hline
R: & We understand that has already commenced, but there is a long way to go. \\
T: & This situation has already commenced, but much still needs to be done. \\
\hline
R: & The restaurant is closed on Sundays. No breakfast is available on Sunday mornings. \\
T: & The restaurant stays closed Sundays so no breakfast is served these days. \\
\hline
R: & Improved central bank policy is another huge factor. \\
T: & Another crucial factor is the improved policy of the central banks. \\
\hline
\end{tabular}
\caption{
Illustrative examples of references (R) paired with back-translations (T).
}
\label{table:NMTexamples}
\end{table}

In this paper, we build NMT systems for three language pairs, then use them to back-translate the non-English side of the training bitext.
The resulting data consists of sentence pairs containing an English reference and the output of an X-to-English NMT system.  Table~\ref{table:NMTexamples} shows examples. 
We use this data for training paraphrastic sentence embeddings, yielding results that are much stronger than when using PPDB and 
competitive with the Simple English Wikipedia data. 

Since bitext is abundant and available for many language pairs and in many data domains, we also develop several methods of filtering the data, including based on sentence length, quality measures, and measures of difference between the reference and its back-translation. We find length to be an effective filtering method, showing that very short length ranges\----where the translation is 1 to 10 words\----are best for learning. 

In studying quality measures for filtering, we train a classifier to predict if a sentence is a reference or a back-translation, then score sentences by the classifier score. 
This investigation allows us to examine the kinds of phenomena that best distinguish NMT output from references in this controlled setting of translating the bitext training data. NMT output has more repetitions of both words and longer $n$-grams, and uses fewer rare words than the references. 

We release our generated sentence pairs to the research community with the hope that the data can inspire others to develop additional filtering methods, to experiment with richer architectures for sentence embeddings, and to further analyze the differences between neural machine translations and references.

\section{Related Work}
We describe related work in learning general-purpose sentence embeddings, work in automatically generating or discovering paraphrases, and finally prior work in leveraging neural machine translation for embedding learning. 

\paragraph{Paraphrastic sentence embeddings.} 
Our learning and evaluation setting is the same as that considered by \newcite{wieting-16-full} and \newcite{wieting2016charagram}, in which the goal is to learn paraphrastic sentence embeddings that can be used for downstream tasks. They trained models on PPDB 
and evaluated them using a suite of semantic textual similarity (STS) tasks and supervised semantic tasks. Others have begun to consider this setting as well~\cite{arora2017simple}. 

Other work in learning general purpose sentence embeddings has used  autoencoders~\cite{SocherEtAl2011:PoolRAE,hill2016learning}, 
encoder-decoder architectures~\cite{kiros2015skip}, or other learning frameworks~\cite{le2014distributed,pham-EtAl:2015:ACL-IJCNLP}. 
\newcite{wieting-16-full} and \newcite{hill2016learning} provide many empirical comparisons to this prior work. For conciseness, we compare only to the strongest configurations from their results. 

\paragraph{Paraphrase generation and discovery.} 
There is a rich history of research in generating or finding naturally-occurring sentential paraphrases~\cite{barzilay2001extracting,dolan2004unsupervised,dolan-05,quirk-04,coster2011simple,Xu:ea:2014,xu2015semeval}. 

The most relevant work uses bilingual corpora, e.g., \newcite{bannard2005paraphrasing}, culminating in PPDB.  
Our goals are highly similar to those of the PPDB project, which has also been produced for many languages~\cite{ganitkevitch2014multilingual} since it only relies on the availability of bilingual text. 

Prior work has shown that PPDB can be used for learning embeddings for words and phrases~\cite{faruqui-15,wieting2015ppdb-short}. However, when learning sentence embeddings, \newcite{wieting-17-full} showed that PPDB is not as effective as sentential paraphrases, especially for recurrent networks. These results are intuitive because the phrases in PPDB are short and often cut across constituent boundaries. 
For sentential paraphrases, \newcite{wieting-17-full} used a dataset developed for text simplification by \newcite{coster2011simple}. It was  created by aligning sentences from Simple English and standard English Wikipedia. We compare our data to both PPDB and this Wikipedia dataset. 

\paragraph{Neural machine translation for paraphrastic embedding learning.} 
\newcite{sutskever2014sequence} trained NMT systems and visualized part of the space of the source language encoder for their English$\rightarrow$French system. 
\newcite{hill2016learning} evaluated the encoders of English-to-X NMT systems as sentence representations, finding them to perform poorly compared to several other methods based on unlabeled data. 
\newcite{E17-1083} adapted trained NMT models to produce sentence similarity scores in semantic evaluations. They 
used pairs of NMT systems, one to translate an English sentence into multiple foreign translations and the other to then translate back to English. 
Other work has used neural MT architectures and training settings to obtain better word embeddings~\cite{hill2014embedding,hill-14}. 

Our approach differs in that we only use the NMT system to generate training data for training sentence embeddings, rather than use it as the source of the model. This permits us to decouple decisions made in designing the NMT architecture from decisions about which models we will use for learning sentence embeddings. Thus we can benefit from orthogonal work in designing neural architectures to embed sentences.

\section{Neural Machine Translation}
 We now describe the NMT systems we use for generating data for learning sentence embeddings. In our experiments, we use three encoder-decoder NMT models: Czech$\rightarrow$English, French$\rightarrow$English, and German$\rightarrow$English. 

We used Groundhog\footnote{Available at \url{https://github.com/sebastien-j/LV_groundhog}.} as
the implementation of the NMT systems for all experiments.  We
generally followed the settings and training procedure from previous
work \cite{bahdanau2014neural,sennrich2015improving}. As such, all
networks have a hidden layer size of~1000 and an embedding layer size
of~620. During training, we used Adadelta \cite{zeiler2012adadelta}, a
minibatch size of~80, and the training set was reshuffled between
epochs. We trained a network for approximately 7 days on a single GPU (TITAN X), then the
embedding layer was fixed and training continued, as suggested by
\newcite{cho2015using}, for 12~hours. Additionally, the softmax was
calculated over a filtered list of candidate translations. Following \newcite{cho2015using}, during decoding, we restrict the softmax layers' output vocabulary to include: the 10000 most common words, the top 25 unigram translations, and the gold translations' unigrams. 

All systems
were trained on the available training data from the WMT15 shared
translation task (15.7 million, 39.2 million, and 4.2 million sentence
pairs for CS$\rightarrow$EN, FR$\rightarrow$EN, and
DE$\rightarrow$EN, respectively). The training data included: Europarl v7 \citep{koehn2005europarl}, the Common Crawl corpus, the UN corpus \cite{eisele2010multiun}, News Commentary v10, the $10^9$ French-English corpus, and CzEng 1.0 \cite{czeng16:2016}. A breakdown of the sizes of these corpora can be found in Table~\ref{tbl:corp}. 
The data was pre-processed using standard pre-processing
scripts found in Moses~\cite{koehn-EtAl:2007:PosterDemo}. Rare words
were split into sub-word units, following
\newcite{sennrich2015neural}. BLEU scores on the WMT2015 test set for each NMT system can be seen in Table~\ref{tbl:bleu}.  

\begin{table}[t]
\setlength{\tabcolsep}{4pt}
\small
\centering
\label{tbl:corp}
\begin{tabular}{|l|r|r|r|}
\cline{2-4}
\multicolumn{1}{c|}{}  & \multicolumn{1}{c|}{Czech} & \multicolumn{1}{c|}{French}           & \multicolumn{1}{c|}{German}         \\ \hline
Europarl               & 650,000 & 2,000,000  & 2,000,000 \\ 
Common Crawl          & 160,000 & 3,000,000  & 2,000,000 \\
News Commentary       & 150,000   & 200,000    & 200,000 \\
UN                    &  \multicolumn{1}{c|}{-} & 12,000,000 & \multicolumn{1}{c|}{-} \\
$10^9$ French-English & \multicolumn{1}{c|}{-} & 22,000,000 & \multicolumn{1}{c|}{-} \\
CzEng                 & 14,700,000 & \multicolumn{1}{c|}{-} & \multicolumn{1}{c|}{-}    \\
\hline
\end{tabular}
\caption{Dataset sizes (numbers of sentence pairs) for data domains used for training NMT systems.}
\end{table}

\begin{table}[t]
\small
\centering
\label{tbl:bleu}
\begin{tabular}{|l|c|}
\hline
Language                   & \% BLEU     \\
\hline
Czech$\rightarrow$English  & 19.7    \\
French$\rightarrow$English & 20.1    \\
German$\rightarrow$English & 28.2    \\
\hline                            
\end{tabular}
\caption{BLEU scores on the WMT2015 test set.}
\end{table}

To produce paraphrases we use ``back-translation'', 
i.e., we use our X$\rightarrow$English NMT systems to translate the non-English sentence in each training sentence pair into English. 
We 
directly use the bitext on which the models were trained. This could potentially lead to pairs in which the reference and translation match exactly, if the model has learned to memorize the reference translations seen during training. However, in practice, since we have so much bitext to draw from, we can easily find data in which they do not match exactly. 

Thus our generated data consists of pairs of English references from the bitext along with the NMT-produced English back-translations. 
We use beam search with a width of 50 to generate multiple translations for each non-English sentence, 
each of which is a candidate paraphrase for the English reference.  

Example outputs of this process are in Table~\ref{table:NMTexamples}, showing some rich paraphrase phenomena in the data. 
These examples show 
non-trivial phrase substitutions (``there is a long way to go'' and ``much still needs to be done''), sentences being merged and simplified, and sentences being rearranged. 
For examples of erroneous paraphrases that can be generated by this process, see Table~\ref{table:classifyanalysis}.

\section{Models and Training}
Our goal is to compare our paraphrase dataset to other datasets by using each to train sentence embeddings, keeping the models and learning procedure fixed. So we select models and a loss function from prior work~\cite{wieting-16-full,wieting-17-full}. 

\subsection{Models}

We wish to embed a word sequence $\sent$ into a fixed-length vector. 
We denote the $t$th word in $\sent$ as $\sentt$, and we denote its word embedding by $x_t$. 
We focus on two models in this paper. 
The first model, which we call \wordavg, simply averages the embeddings $x_t$ of all words in $\sent$. The only parameters learned in this model are those in the word embeddings themselves, which are stored in the word embedding matrix $W_w$. 
This model was found by \newcite{wieting-16-full} to perform very strongly for semantic similarity tasks. 

The second model, the \ganlong (\gan)~\cite{wieting-17-full}, combines the benefits of \wordavg and long short-term memory (LSTM) recurrent neural networks~\cite{hochreiter1997long}. It first uses an LSTM to generate a hidden vector, $h_t$, for each word $\sentt$ in $\sent$. Then $h_t$ is used to compute a gate that is elementwise-multiplied with $x_t$, resulting in a new hidden vector $a_t$ for each step $t$:
\begin{equation}
a_t = x_t\odot\sigma(W_xx_t + W_hh_t + b) \label{eq:gan}
\end{equation}
\noindent 
where $W_x$ and $W_h$ are parameter matrices, $b$ is a parameter vector, and $\sigma$ is the elementwise logistic sigmoid function. After all $a_t$ have been generated for a sentence, they are averaged to produce the embedding for that sentence. The \gan reduces to \wordavg if the output of the gate is always 1. This model includes as learnable parameters those of the LSTM, the word embeddings, and the additional parameters in Eq.~\eqref{eq:gan}. We use $W_c$ to denote the ``compositional'' parameters, i.e., all parameters other than the word embeddings.

\subsection{Training}

We follow the training procedure of \newcite{wieting2015ppdb-short} and \newcite{wieting-16-full}. 
The training data is a set $S$ of paraphrastic  
pairs $\langle s_1, s_2\rangle$ and 
we optimize a margin-based loss: 

\begin{small}
\begin{align}
&\underset{W_c,W_w}{\text{min}} \frac{1}{|S|}\Bigg(\sum_{\langle s_1,s_2\rangle \in S} 
\max(0,\delta - \cos(g(s_1), g(s_2))\\
&+ \cos(g(s_1), g(t_1))) + \max(0,\delta - \cos(g(s_1),g(s_2))\\
&+ \cos(g(s_2), g(t_2)))\bigg) + \lambda_c\norm{W_c}^2  + \lambda_{w}\norm{W_{w_{\mathit{initial}}} - W_w}^2
\label{eq:obj}
\end{align} 
\end{small}

\noindent where $g$ is the model (\wordavg or \gan), $\delta$ is the margin, $\lambda_{c}$ and $\lambda_{w}$ are regularization parameters, $W_{w_{\mathit{initial}}}$ is the initial word embedding matrix, and $t_1$ and $t_2$ are ``negative examples'' taken from a mini-batch during optimization. 
The intuition is that we want the two texts to be more similar to each other ($\cos(g(s_1), g(s_2))$) than either is to their respective negative examples $t_1$ and $t_2$, by a margin of at least $\delta$. 
To select $t_1$ and $t_2$, 
we choose the most similar sentence in some set (other than those in the given pair). 
For simplicity 
we use the mini-batch for this set, i.e., 
we choose $t_1$ for a given $\langle s_1, s_2\rangle$ as follows:
\begin{equation}
t_1 = \argmax_{t : \langle t, \cdot\rangle \in S_b \setminus \{\langle s_1, s_2\rangle\}} \cos(g(s_1), g(t))
\end{equation}
\noindent where $S_b\subseteq S$ is the current mini-batch. 
That is, we want to choose a negative example $t_i$ that is similar to $s_i$ according to the current model. 
The downside is that we may occasionally choose a phrase $t_i$ that is actually a true paraphrase of $s_i$.

\section{Experiments}  \label{sec:exp}
We now investigate how best to use our generated paraphrase data for training universal paraphrastic sentence embeddings. We consider 10 data sources: Common Crawl (CC), Europarl (EP), and News Commentary (News) from all 3 language pairs, as well as the $10^9$ French-English data (Giga). 
We extract 150,000 reference/back-translation pairs from each data source. We use 100,000 of these to mine for training data for our sentence embedding models, and the remaining 50,000 are used as 
train/validation/test data for the reference classification and language models described below. 

\subsection{Evaluation}

We evaluate the quality of a paraphrase dataset 
by using the experimental setting of \newcite{wieting-16-full}. We use the paraphrases as training data to create paraphrastic sentence embeddings, then evaluate the embeddings on the 
SemEval semantic textual similarity (STS) tasks from 2012 to 2015~\cite{agirre2012semeval,diab2013eneko,agirre2014semeval,agirre2015semeval}, the SemEval 2015 Twitter task \cite{xu2015semeval}, and the SemEval 2014 SICK Semantic Relatedness task \cite{marelli2014semeval}. 

Given two sentences, the aim of the STS tasks is to predict their similarity on a 0-5 scale, where 0 indicates the sentences are on different topics and 5 indicates that they are completely equivalent. 
As our test set, we report the average Pearson's $r$ over these 22 sentence similarity tasks. 
As development data, we use the 2016 STS tasks~\cite{agirre2016semeval},  
where the tuning criterion is the average Pearson's $r$ over its 5 datasets. 

\subsection{Experimental Setup}

For fair comparison among different datasets and dataset filtering methods described below, we use only 24,000 training examples for nearly all experiments. Different filtering methods
produce different amounts of training data, and using 24,000 examples allows us to keep the amount of training data constant across filtering methods. It also allows us to complete these several thousand experiments in a reasonable amount of time. In Section~\ref{sec:scaling} below, we discuss experiments that scale up to larger amounts of training data. 

We use \paragramsl embeddings~\cite{wieting2015ppdb-short} to initialize the word embedding matrix ($W_w$) for both models. For all experiments, we fix the mini-batch size to 100, $\lambda_{w}$ to 0, $\lambda_{c}$ to 0, and the margin $\delta$ to 0.4. We train \wordavg for 20 epochs, and the \gan for 3, since it converges much faster. For optimization we use Adam~\cite{kingma2014adam} with a learning rate of 0.001. 

We compare to two data resources used in previous work to learn paraphrastic sentence embeddings. 
The first is phrase pairs from PPDB, used by \newcite{wieting-16-full} and \newcite{wieting2016charagram}. PPDB comes in different sizes (S, M, L, XL, XXL, and XXXL), where each larger size subsumes all smaller ones. The pairs in PPDB are sorted by a confidence measure and so the smaller sets contain higher precision paraphrases. We use PPDB XL in this paper, which consists of fairly high precision paraphrases.
The other data source is the aligned Simple English / standard English Wikipedia data developed by \newcite{coster2011simple} and used for learning paraphrastic sentence embeddings by \newcite{wieting-17-full}. We refer to this data source as ``\simpwiki''. We refer to our back-translated data as ``NMT''. 

\subsection{Dataset Comparison}

We first compare datasets, randomly sampling 24,000 sentence pairs from each of PPDB, \simpwiki, and each of our NMT datasets. The only hyperparameter to tune for this experiment is the stopping epoch, which we tune based on our development set. The results are shown in Table~\ref{table:randomresults}. 

\begin{table}
\setlength{\tabcolsep}{4pt}
\small
\centering
\begin{tabular} { | l | l || c | c |} 
\hline
 Lang. & Data & \gan & \wordavg \\
\hline
 \multicolumn{2}{|c||}{\simpwiki} & 67.2 & 65.8 \\
 \multicolumn{2}{|c||}{PPDB}      & 64.5 & 65.8 \\
 \hline
    & CC & 65.5 & 65.4 \\
 CS & EP & 66.5 & 65.1 \\
    & News & 67.2 & 65.1 \\
 \hline
    & CC & 67.3 & 66.1 \\
 \multirow{2}{*}{FR} & EP & \bf 67.8 & 65.7 \\
    & Giga & 67.4 & 65.9 \\
    & News & 67.0 & 65.2 \\
 \hline
    & CC & 66.5 & \bf 66.2 \\
 DE & EP & 67.2 & 65.6 \\
    & News & 66.5 & 64.7 \\
\hline
\end{tabular}
\caption{\label{table:randomresults}
Test results (average Pearson's $r\times 100$ over 22 STS datasets) using a random selection of 24,000 examples from each data source considered in this paper. 
}
\end{table}

We find that the NMT datasets are all effective as training data, outperforming PPDB in all cases when using the \gan. There are exceptions when using \wordavg, for which PPDB is quite strong. This is sensible because \wordavg is not sensitive to word order, so the fragments in PPDB do not cause problems. However, when using the \gan, which is sensitive to word order, the NMT data is consistently better than PPDB. It often exceeds the performance of training on the \simpwiki data, which consists entirely of human-written sentences. 

\subsection{Filtering Methods}
Above we showed that the NMT data is better than PPDB when using a \gan and often as good as \simpwiki. Since we have access to so much more NMT data than SimpWiki (which is limited to fewer than 200k sentence pairs), we next experiment with several approaches for filtering the NMT data. 
We first consider filtering based on length, described in Section~\ref{sec:length}. 
We then consider filtering based on several quality measures designed to find more natural and higher-quality translations, described in Section~\ref{sec:quality}. 
Finally, we consider several measures of diversity. By diversity we mean here a measure of the lexical and syntactic difference between the reference and its paraphrase. 
We describe these experiments in Section~\ref{sec:diversity}. 
We note that these filtering methods are not all mutually exclusive and could be combined, though in this paper we experiment with each individually and leave combination to future work. 

\subsection{Length Filtering}
\label{sec:length}

We first consider filtering candidate sentence pairs by length, i.e., the number of tokens in the translation. The tunable parameters are the upper and lower bounds of the translation lengths. 

We experiment with a partition of length ranges, showing the results in Table~\ref{table:length-bins}. These results are averages across all language pairs and data sources of training data for each length range shown. We find it best to select NMT data where the translations have between 0 and 10 tokens, with performance dropping as sentence length increases. This is true for both the \gan and \wordavg models. We do the same filtering for the \simpwiki data, though the trend is not nearly as strong. This may be because machine translation quality drops as sentence length increases. This trend appears even though the datasets with higher ranges have more tokens of training data, since only the number of training sentence pairs is kept constant across configurations. 

We then tune the length range using our development data, considering the following length ranges: [0,10], [0,15], [0,20], [0,30], [0,100], [10,20], [10,30], [10,100], [15,25], [15,30], [15,100], [20,30], [20,100], [30,100]. 
We tune over ranges as well as language, data source, and stopping epoch, each time training on 24,000 sentence pairs. 
We report the average test results over all languages and datasets in Table~\ref{table:length-results-gan-wa}. We compare to a baseline that draws a random set of data, showing that length-based filtering leads to gains of nearly half a point on average across our test sets. The tuned length ranges are short for both NMT and \simpwiki. 

The distribution of lengths in the NMT and \simpwiki data is fairly similar. The 10 NMT datasets all have mean translation lengths between 22 and 28 tokens, with common crawl on the lower end
and Europarl on the higher end.
The data has fairly large standard deviations (11-25 tokens) indicating that there are some very long translations in the data. \simpwiki has a mean length of 24.2 and a standard deviation of 13.1 tokens.

\begin{table}[t]
\small
\centering
\begin{tabular} { | l | l | c | c | c | c |} 
\cline{3-6}
\multicolumn{2}{c|}{} & \multicolumn{4}{c|}{Length Range} \\
\hline
 Data & Model & 0-10 & 10-20 & 20-30 & 30-100 \\
\hline
 \multirow{2}{*}{\simpwiki} & \gan & 67.4 & 67.7 & 67.1 & 67.3 \\
                       & \wordavg & 65.9 & 65.7 & 65.6 & 65.9 \\
 \hline
 \multirow{2}{*}{NMT} & \gan & 66.6 & 66.5 & 66.0 & 64.8 \\
                      & \wordavg & 65.7 & 65.6 & 65.3 & 65.0 \\
\hline
\end{tabular}
\caption{\label{table:length-bins}
Test correlations for our models when trained on sentences with particular length ranges (averaged over languages and data sources for the NMT rows). Results are on STS datasets (Pearson's $r \times 100$). 
}
\end{table}

\begin{table}
\small
\centering
\begin{tabular} { | l | c | c | c | c |} 
\cline{2-5}
\multicolumn{1}{c|}{} & \multicolumn{2}{c|}{NMT} & \multicolumn{2}{c|}{\simpwiki} \\
\hline
Filtering Method & \gan  & \wordavg & \gan & \wordavg \\
\hline
None (Random) & 66.9 & 65.5 & 67.2 & 65.8\\
\hline
Length & 67.3 & 66.0 & 67.4 & 66.2 \\
Tuned Len. Range & [0,10] & [0,10] & [0,10] & [0,15]\\
\hline
\end{tabular}
\caption{\label{table:length-results-gan-wa}
Length filtering test results after tuning length ranges on development data (averaged over languages and data sources for the NMT rows). Results are on STS datasets (Pearson's $r \times 100$). 
}
\end{table}

\subsection{Quality Filtering}
\label{sec:quality}
We also consider filtering based on several measures of the ``quality'' of the back-translation: 
\begin{itemizesquish}
\item {\bf Translation Cost}: We use the cost (negative log likelihood) of the translation from the NMT system, divided by the number of tokens in the translation. 
\item {\bf Language Model}: We train a separate language model for each language/data pair on 40,000 references that are separate from the 100,000 used for mining data. Due to the small data size, we train a 3-gram language model and use the KenLM toolkit~\cite{heafield2011kenlm}. 
\item {\bf Reference/Translation Classification}: We train binary classifiers to predict whether a given sentence is a reference or translation (described in Section~\ref{sec:classification}). 
We use the probability of being a reference as the score for filtering. 
\end{itemizesquish}

For translation cost, we tune the upper bound of the cost 
over the range $[0.2, 1]$ using increments of 0.1. 
For the language model, we tune an upper bound on the perplexity of the  translations among the set $\{25,50,75,100,150,200,\infty\}$. For the classifier, we tune the minimum probability of being a reference over the range $[0, 0.9]$ using increments of 0.1. 

\begin{table}
\small
\centering
\begin{tabular} { | l || c | c |} 
\hline
Filtering Method & \gan  & \wordavg \\
\hline
None (Random) & 66.9 & 65.5 \\
\hline
Translation Cost & 66.6 & 65.4\\
Language Model & 66.7 & 65.5\\
Reference Classification & 67.0 & 65.5\\
\hline
\end{tabular}
\caption{\label{table:qua-results-gan-wa}
Quality filtering test results after tuning quality hyperparameters on development data (averaged over languages and data sources for the NMT rows). Results are on STS datasets (Pearson's $r \times 100$).
}
\end{table}

Table~\ref{table:qua-results-gan-wa} shows average test results over all languages and datasets after tuning hyperparameters on our development data for each. The translation cost and language model are not helpful for filtering, as random selection outperforms them. Both methods are outperformed by the reference classifier, which slightly outperforms random selection when using the stronger \gan model. 
We now discuss further how we trained the reference classifier and the data characteristics that it reveals. 
We did not experiment with quality filtering for \simpwiki since it is human-written text.

\subsubsection{Reference/Translation Classification}
\label{sec:classification}

We experiment with predicting whether a given sentence is a reference or a back-translation. We train two kinds of binary classifiers, 
one using an LSTM and the other using word averaging, followed by a softmax layer. 
We select 40,000 reference/translation pairs for training and 5,000 for each of  validation and testing. A single example is a sentence with label 1 if it is a reference translation and 0 if it is a translation. 

In training, we consider the entire $k$-best list as examples of translations, selecting one translation to be the 0-labeled example. We either do this randomly or 
we score each sentence in the $k$-best list using our model and select the one with the highest probability of being a reference as the 0-labeled example. We tune this choice as well as an $L_2$ regularizer on the word embeddings (tuned over $\{10^{-5},10^{-6},10^{-7},10^{-8},0\}$). We use \paragramsl embeddings~\cite{wieting2015ppdb-short} to initialize the word embeddings for both models. 
Models were trained by minimizing cross entropy for 10 epochs using Adam 
with learning rate 0.001. We performed this procedure separately for each of the 10 language/data pairs. 

\begin{table}
\setlength{\tabcolsep}{4pt}
\small
\centering
\begin{tabular} { | l | l | l || c | c | c |} 
\hline
Model & Lang. & Data & Test Acc. & + Acc. & - Acc. \\
\hline
 &  & CC & 72.2 & 72.2 & 72.3 \\
 & CS & EP & 72.3 & 64.3 & 80.3 \\
 &  & News & 79.7 & 73.2 & 86.3 \\
\cline{2-6}
 &  & CC & 80.7 & 82.1 & 79.3 \\
\multirow{2}{*}{LSTM} & \multirow{2}{*}{FR} & EP & 79.3 & 75.2 & 83.4 \\
 &  & Giga & \bf 93.1 & \bf 92.3 & 93.8 \\
 &  & News & 84.2 & 81.2 & 87.3 \\
\cline{2-6}
 &  & CC & 79.3 & 71.7 & 86.9 \\
 & DE & EP & 85.1 & 78.0 & 92.2 \\
 &  & News & 89.8 & 82.3 & \bf 97.4 \\
\hline
 &  & CC & 71.2 & 68.9 & 73.5 \\
 & CS & EP & 69.1 & 63.0 & 75.1 \\
 &  & News & 77.6 & 71.7 & 83.6 \\
\cline{2-6}
 &  & CC & 78.8 & 80.4 & 77.2 \\
\multirow{2}{*}{\wordavg} & \multirow{2}{*}{FR} & EP & 78.9 & 75.5 & 82.3 \\
 &  & Giga & \bf 92.5 & \bf 91.5 & 93.4 \\
 &  & News & 82.8 & 81.1 & 84.5 \\
\cline{2-6}
 &  & CC & 77.3 & 70.4 & 84.1 \\
 & DE & EP & 82.7 & 73.4 & 91.9 \\
 &  & News & 87.6 & 80.0 & \bf 95.3 \\
\hline
\end{tabular}
\caption{\label{table:classificationresults}
Results of reference/translation classification (accuracy$\times 100$). The highest score in each column is in boldface. Final two columns show accuracies of positive (reference) and negative classes, respectively. 
}
\end{table}

The results are shown in Table~\ref{table:classificationresults}. While performance varies greatly across data sources, the LSTM always outperforms the word averaging model.  
For our translation-reference classification, we note that our results can be further improved.  We also trained models on 90,000 examples, essentially doubling the amount of data, and the results improved by about 2\% absolute on each dataset on both the validation and testing data. 

\paragraph{Analyzing Reference Classification.}

We inspected the output of our reference classifier and noted a few qualitative trends which we then verified empirically. First, neural MT systems tend to use a smaller vocabulary and exhibit more restricted use of phrases. 
They correspondingly tend to show more repetition in terms of both words and longer $n$-grams. This hypothesis can be verified empirically in several ways. We do so by calculating the entropy of the unigrams and trigrams for both the references and the translations from our 150,000 reference-translation pairs.\footnote{We randomly selected translations from the beam search.} We also calculate the repetition percentage of unigrams and trigrams in both the references and translations. This is defined as the percentage of words that are repetitions (i.e., have already appeared in the sentence). For unigrams, we only consider words consisting of at least 3 characters.

The results are shown in Table~\ref{table:repetitionentropy}, in which we 
subtract the translation value from the reference value for each measure. The translated text has lower $n$-gram entropies and higher rates of repetition. 
This appears for all datasets, but is strongest for common crawl and French-English $10^9$. 

\begin{table}
\setlength{\tabcolsep}{4pt}
\small
\centering
\begin{tabular} { | l | l || c | c | c | c |} 
\hline
 Lang. & Data & Ent. (uni) & Ent. (tri) & Rep. (uni) & Rep. (tri) \\
 \hline
  & CC & 0.50 & 1.13 & -7.57\% & -5.58\% \\
 CS & EP & 0.14 & 0.31 & -0.88\% & -0.11\% \\
  & News & 0.16 & 0.31 & -0.96\% & -0.16\% \\
 \hline
 & CC & 0.97 & 1.40 & -8.50\% & -7.53\% \\
\multirow{2}{*}{FR} & EP & 0.51 & 0.69 & -1.85\% & -0.58\% \\
  & Giga & 0.97 & 1.21 & -5.30\% & -7.74\% \\
  & News & 0.67 & 0.75 & -2.98\% & -0.85\% \\
 \hline
  & CC & 0.29 & 0.57 & -1.09\% & -0.73\% \\
 DE & EP & 0.32 & 0.53 & -0.14\% & -0.11\% \\
  & News & 0.40 & 0.37 & -1.02\% & -0.24\% \\
 \hline
 \multicolumn{2}{|c||}{All} & 0.46 & 0.74 & -2.80\% & -2.26\% \\
\hline
\end{tabular}
\caption{\label{table:repetitionentropy}
Differences in entropy and repetition of unigrams/trigrams in references and translations. Negative values indicate translations have a higher value, so references show consistently higher entropies and 
lower repetition rates. 
}
\end{table}

We also noticed that translations are less likely to use rare words, instead willing to use a larger sequence of short words to convey the same meaning. 
We found that translations were sometimes more vague and, unsurprisingly, were more likely to be ungrammatical. 

We check whether our classifier is learning these patterns by computing the reference probabilities $P(\text{R})$ of 100,000 randomly sampled translation-reference pairs from each dataset (the same used to train models). We then compute the correlation between our classification score and different metrics: the repetition rate of the sentence, the average inverse-document frequency (IDF) of the sentence,\footnote{Wikipedia was used to calculate the frequencies of the tokens. All tokens were lowercased.} and the length of the translation. 

\begin{table}
\setlength{\tabcolsep}{4pt}
\small
\centering
\begin{tabular} { | l | c |} 
\hline
 Metric & Spearman's $\rho$ \\
\hline
Unigram repetition rate  & -35.1 \\
Trigram repetition rate & -18.4 \\
Average IDF & \phantom{-}27.8 \\
Length & -34.0 \\
\hline
\end{tabular}
\caption{\label{table:correlations}
Spearman's $\rho$ between our reference classifier probability and various measures.
}
\end{table}

\begin{table}[t]
\setlength{\tabcolsep}{3pt} 
\small
\centering
\begin{tabular} { | c p{0.76\columnwidth} c| } \hline
& Sentence & $P(\text{R})$ \\
\hline
R: & Room was comfortable and the staff at the front desk were very helpful. & 1.0 \\
T: & The staff were very nice and the room was very nice and the staff were very nice. & $<\!0.01$ \\ 
\hline
R: & The enchantment of your wedding day, captured in images by Flore-Ael Surun. & 0.98 \\
T: & The wedding of the wedding, put into images by Flore-Ael A. & $<\!0.01$ \\ %9.8e-3 \\
\hline
R: & Mexico and Sweden are longstanding supporters of the CTBT. & 1.0 \\
T: & Mexico and Sweden have been supporters of CTBT for a long time now. & 0.06 \\
\hline
R: & We thought Mr Haider ' s Austria was endangering our freedom. & 1.0 \\
T: & We thought that our freedom was put at risk by Austria by Mr Haider. & 0.09 \\
\hline
\end{tabular}
\caption{
Illustrative examples of references (R) and back-translations (T), 
along with probabilities from the reference classifier. See text for details. 
}
\label{table:classifyanalysis}
\end{table}

The results are shown in Table~\ref{table:correlations}. 
Negative correlations with repetitions indicates that fewer repetitions lead to higher $P(\text{R})$. A positive correlation with average IDF indicates that $P(\text{R})$ rewards the use of rare words. Interestingly, negative correlation with length suggests that the classifier prefers 
more concise sentences.\footnote{This is noteworthy because the average sentence length of translations and references is not significantly different.}
We show examples of these phenomena 
in Table~\ref{table:classifyanalysis}. 
The first two examples show the tendency of NMT to repeat words and phrases. The second two show how they tend to use sequences of common words (``put at risk'') rather than rare words (``endangering'').

\subsection{Diversity Filtering}
\label{sec:diversity}

\begin{table}
\setlength{\tabcolsep}{4pt}
\small
\centering
\begin{tabular} { | l | c | c | c | c |} 
\cline{2-5}
\multicolumn{1}{c|}{} & \multicolumn{2}{c|}{NMT} & \multicolumn{2}{c|}{\simpwiki} \\
\hline
Filtering Method & \gan & \wordavg & \gan & \wordavg \\
\hline
Random & 66.9 & 65.5 & 67.2 & 65.8\\
\hline
Unigram Overlap & 66.6 & 66.1 & 67.8 & 67.4 \\
Bigram Overlap & 67.0 & 65.5 & 68.0 & 67.2 \\
Trigram Overlap & 66.9 & 65.4 & 67.8 & 66.6 \\
BLEU Score  & 67.1 & 65.3 & 67.5 & 66.5 \\
\hline
\end{tabular}
\caption{\label{table:diversity-results}
Diversity filtering test results after tuning filtering hyperparameters on development data (averaged over languages and data sources for the NMT rows). Results are on STS datasets (Pearson's $r \times 100$).
}
\end{table}

We consider several filtering criteria based on measures that encourage particular amounts of disparity between the reference and its back-translation:
\begin{itemizesquish}
\item {\bf $n$-gram Overlap}: Our $n$-gram overlap measures are calculated by counting $n$-grams of a given order in both the reference and translation, then dividing the number of shared $n$-grams by the total number of $n$-grams in the reference or translation, whichever has fewer. We use three $n$-gram overlap scores ($n\in\{1,2,3\}$). 
\item {\bf BLEU Score}: We use a smoothed sentence-level BLEU variant from \newcite{nakov-guzman-vogel:2012:PAPERS} that uses smoothing for all $n$-gram lengths and also smooths the brevity penalty. 
\end{itemizesquish}

For both methods, the tunable hyperparameters are the upper and lower bounds for the above scores.
We tune over the cross product of lower bounds $\{0,0.1,0.2,0.3\}$ and upper bounds $\{0.6,0.7,0.8,0.9,1.0\}$. 
Our intuition is that the best data will have some amount of $n$-gram overlap, but not too much. Too much $n$-gram overlap will lead to pairs that are not useful for learning. 

The results are shown in Table~\ref{table:diversity-results}, for both models and for both NMT and \simpwiki. We find that the diversity filtering methods lead to consistent improvements when training on \simpwiki. We believe this is because many of the sentence pairs in \simpwiki are near-duplicates and these filtering methods favor data 
with more differences. 

Diversity filtering can also help when selecting NMT data, though the differences are smaller. We do note that unigram overlap is the strongest filtering strategy for \wordavg. When looking at the threshold tuning, the best lower bounds are often 0 or 0.1 and the best upper bounds are typically 0.6-0.7, indicating that sentence pairs with a high degree of word overlap are not useful for training. We also find that the \gan benefits more from filtering based on higher-order $n$-gram overlap than \wordavg. 

\subsection{Scaling Up}
\label{sec:scaling}

Unlike the \simpwiki data, which is naturally limited and only available for English, we can scale our approach. Since we use data on which the NMT systems were trained and perform back-translation, we can easily produce large training sets of paraphrastic sentence pairs for many languages and data domains, limited only by the availability of bitext. 

\begin{table}
\small
\centering
\begin{tabular} { | l || c | c |} 
\hline
Data & \gan  & \wordavg \\
\hline
PPDB & 64.6 & 66.3\\
\simpwiki (100k/168k) & 67.4 & \bf 67.7 \\
\hline
CC-CS (24k) & 66.8 & - \\
CC-DE (24k) & - & 66.6 \\
\hline
CC-CS (100k) & \bf 68.5 & - \\
CC-DE (168k) & - & 67.6\\
\hline
\end{tabular}
\caption{\label{table:scalingup}
Test results when using more training data. More data helps both \wordavg and \gan, where both models get close to or surpass training on \simpwiki and comfortably surpass the PPDB baseline. The amount of training examples used is in parentheses.
}
\end{table}

To test this, we took the tuned filtering approaches and language/data pairs (according to our development dataset only
), and trained them on more data. These were the CC-CS for \gan and CC-DE for \wordavg models. We trained them with the same amount of sentence pairs from \simpwiki.\footnote{Since the CC-CS data was the smallest dataset used to train the CS NMT system (See Table~\ref{tbl:corp}) and limited, we only used 100,000 pairs for the \gan experiment. For \wordavg, we used the full 167,689.} We also compare to PPDB XL, and since PPDB has fewer tokens per example, we use enough PPDB data so that it has at least as many tokens as the \simpwiki data used in the experiment.\footnote{This was 800,011 and 1,341,188 pairs in the \gan and \wordavg respectively.} 

The results are shown in Table~\ref{table:scalingup}, where the NMT data surpasses \simpwiki, while both \simpwiki and the NMT data perform similarly for \wordavg. PPDB is soundly beaten by both data sources. Moreover, the CC-CS and CC-DE data have greatly improved their performance from training on just 24,000 examples from 66.77 and 66.61 respectively, providing more evidence that this approach does indeed scale.

\section{Conclusion}
We have shown how back-translating can be an effective paraphrase-generation technique, by using it to produce paraphrastic embeddings at least on par with human-written paraphrase datasets and significantly surpassing PPDB. We have also shown that filtering can improve the generated paraphrase corpus, and we explored a variety of filtering techniques. In doing so, we also identified characteristics that distinguish NMT output from human-written sentences. Future work will aim to improve this generation process, for instance by devising more sophisticated filtering techniques.

\section*{Acknowledgments}

This research used resources of the Argonne Leadership Computing Facility, which is a DOE Office of Science User Facility supported under Contract DE-AC02-06CH11357. We thank the developers of Theano~\cite{2016arXiv160502688short} and NVIDIA Corporation for donating GPUs used in this research.

\bibliography{emnlp2017}
\bibliographystyle{emnlp_natbib}

\end{document}